\documentclass{article} 
\usepackage{graphicx}
\usepackage{subcaption}
\usepackage{amsmath}
\usepackage{amssymb}
\usepackage{float}
\usepackage{url}
\usepackage{xcolor}
\usepackage{makecell}
\usepackage{cite}
\usepackage{wrapfig}
\makeatletter
\let\NAT@parse\undefined
\makeatother
\usepackage{hyperref}

\newcommand{\eg}{\emph{e.g.}, }
\newcommand{\ie}{\emph{i.e.}, } 

\graphicspath{ {images/} }

\title{\LARGE \bf Continuous-Time Spline Visual-Inertial Odometry}

\author{Jiawei Mo$^{1}$ and Junaed Sattar$^{2}$
\thanks{The authors are with the Department of Computer Science and Engineering, Minnesota Robotics Institute, University of Minnesota Twin Cities, Minneapolis, MN, USA.
{\tt\small \{$^{1}$moxxx066, $^{2}$junaed\} at umn.edu.}}
}
\date{}

\begin{document}

\maketitle
\thispagestyle{empty}
\pagestyle{empty}

\begin{abstract}
We propose a continuous-time spline-based formulation for visual-inertial odometry (VIO). Specifically, we model the poses as a cubic spline, whose temporal derivatives are used to synthesize linear acceleration and angular velocity, which are compared to the measurements from the inertial measurement unit (IMU) for optimal state estimation. The spline boundary conditions create constraints between the camera and the IMU, with which we formulate VIO as a constrained nonlinear optimization problem. Continuous-time pose representation makes it possible to address many VIO challenges, \eg rolling shutter distortion and sensors that may lack synchronization. We conduct experiments on two publicly available datasets that demonstrate the state-of-the-art accuracy and real-time computational efficiency of our method.
\end{abstract}

\section{Introduction}
\label{sec:introduction}
Simultaneous localization and mapping (SLAM) has been an active research area over the past few decades~\cite{cadena2016past,thrun2007simultaneous}, with the goal of estimating robot pose using various sensors (\eg laser range-finder, GPS, cameras). Fusion of measurements from multiple sensors combines their advantages and thus improves accuracy and robustness. Due to the ubiquity and energy-efficiency of cameras and inertial measurement units (IMU) (\eg smartphones), visual-inertial odometry (VIO)~\cite{delmerico2018bench} has become a popular sensor fusion approach. Camera and IMU are complementary sensors for SLAM: a camera estimates robot motion up to an unknown scale while an IMU estimates the metric scale; additionally, the IMU estimates inter-frame motion for the visual system to improve accuracy and robustness.

\begin{figure}[ht]
    \centering
    \includegraphics[width=\textwidth]{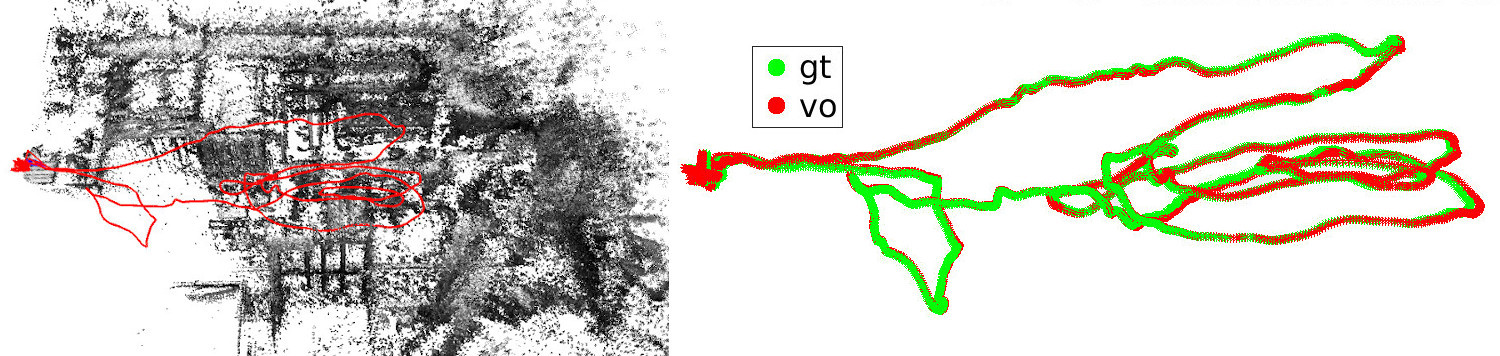}
    \caption{The proposed VIO system running on MH1 of EuRoC dataset. The estimated trajectory (red) is well aligned (by SE(3)) to the ground-truth poses (green). The scale is also recovered accurately.}
    \label{fig:mh1}
\end{figure}

Early work in VIO is mostly dominated by \textit{filtering-based} approaches~\cite{mourikis2007multi, li2012impro, bloesch2017iterated}. Many of these algorithms solve the sensor fusion problem based on extended Kalman filter (EKF)~\cite{thrun2002probabilistic}, where IMU measurements are used for propagation and camera frames are used for the update. A well-established technique is MSCKF~\cite{mourikis2007multi, li2012impro}. It estimates the orientation, position, and velocity of a robot. The IMU measures the angular velocity and linear acceleration in its local frame. The angular velocity measurements between camera frames are integrated to propagate the orientation. However, since the position and velocity propagation depend on orientation, which also changes simultaneously, the Runge-Kutta method~\cite{butcher1996history} is used for approximating it, making the integration process computationally expensive. ROVIO~\cite{bloesch2017iterated} is another development of filtering-based VIO approach. Its propagation process uses a simplified (by using the mean of IMU measurements) IMU preintegration approach~\cite{forster2015imu} to improve computational efficiency. ROVIO also differs from MSCKF with respect to how they process visual input. MSCKF tracks feature across frames and minimizes reprojection error, while ROVIO minimizes photometric error directly without feature extraction and matching. We usually refer to the former process as \textit{feature-based} approach and the latter one as \textit{direct} approach. Direct approaches exhibit higher accuracy and robustness especially in environments with less texture or with repetitive textures~\cite{forster2014svo, engel2017direct}.

Although filtering-based approaches are computationally efficient, the linearization error from EKF may lead to drift and inconsistency~\cite{huang2011obser}. In the past decade, \textit{optimization-based} VIO approaches~\cite{leutenegger2015keyframe, mur2017visual, qin2018vins, von2018direct} have been proposed to solve the linearization error. These methods usually integrate consecutive IMU measurements as a unit constraint. For optimization-based approaches, the optimal state is estimated by iteratively solving nonlinear equations. Once the state is updated, the IMU measurement needs to be re-integrated with the new pose, IMU bias, etc., which is highly computationally expensive. For computational efficiency, most recent work (\eg \cite{qin2018vins, von2018direct}) adopts the IMU preintegration approach~\cite{forster2015imu}, which integrates IMU measurements in the local frame and uses linearized bias in its error function in order to avoid re-integration.

In addition to the methods discussed above that are discrete-time, several continuous-time VIO approaches (\eg \cite{lovegrove2013spline, kerl2015dense, mueggler2018continuous}) have been proposed. The continuous-time pose representation can be useful for many VIO-related challenges (\eg rolling shutter distortion~\cite{lovegrove2013spline}) and tasks (\eg smooth trajectory planning). Typically, they adopt the cumulative B-spline~\cite{qin2000general} to represent the continuous pose. The spline pose representation being temporally differentiable enables us to compute analytical time derivatives and synthesize IMU measurements, which are compared to the real IMU measurements for the purpose of sensor fusion. Compared to conventional spline, B-spline is favored due to its locality~\cite{qin2000general}; additionally, there is no need to explicitly maintain the spline boundary conditions. However, B-spline calculation is time consuming~\cite{sommer2020efficient}, and the VIO systems based on B-spline can barely run in real-time~\cite{lovegrove2013spline}. Improvements have been proposed in \cite{sommer2020efficient}, but B-spline calculation is still slow. Consequently, B-spline pose representation finds use in sensor calibration~\cite{furgale2012continuous, lovegrove2013spline} without real-time requirements.

In this paper, we propose a novel VIO approach that combines the advantages of optimization-based direct approaches and spline pose representation. On the vision side, we adopt the state-of-the-art direct visual odometry DSO~\cite{engel2017direct} to provide exteroceptive pose estimation by minimizing photometric error directly. The major novelty is that we use conventional spline rather than the B-spline. Compared to B-spline, where the control knots are abstract and do not pass through the trajectory, the conventional spline representation is more straightforward. Specifically, we use a cubic polynomial function of time to represent a spline segment between each pair of consecutive DSO keyframes. The constant term of the polynomial is exactly the keyframe pose; the higher-order terms are the temporal derivatives of the spline which are used to synthesize IMU readings and generate the IMU error term for optimization. The core of this work is that we formulate VIO as a \textit{constrained nonlinear optimization problem}. The constraints are the spline boundary conditions (see Sec.~\ref{sec:spline_intro}), which create the constraints between the camera and the IMU. Too many constraints will significantly slow down the system. However, we can achieve real-time performance because DSO maintains a sliding window of keyframes (typically $\leq 7$) for optimization, making the spline scale relatively small and spline boundary conditions manageable. Overall, our contributions are: i) we propose a method that uses the (conventional) cubic spline to represent continuous poses and formulate the VIO as a constrained nonlinear optimization problem; ii) our experiments on two publicly available datasets demonstrate its state-of-the-art performance; iii) we make our implementation publicly available\footnote{\url{https://github.com/IRVLab/spline_vio}}. Fig.~\ref{fig:mh1} shows the results of our VIO system on MH1 of EuRoC dataset~\cite{burri2016euroc}, where the trajectory is well aligned to the ground-truth poses and the scale is recovered accurately.
\section{Methodology}
\label{sec:methodology}
The outline of the proposed approach is that we fit a spline to the poses in DSO and use its temporal derivatives for IMU fusion. We briefly introduce DSO and spline representation; then we discuss how they are used in the proposed method to solve the VIO problem.

\subsection{Notation}
\begin{itemize}
    \item $w$: world coordinate frame
    \item $i$: IMU coordinate frame
    \item $c$: camera coordinate frame
    \item bold lower-case letters (\eg $\mathbf{a}$): vectors
    \item bold upper-case letters (\eg $\mathbf{A}$): matrices
    \item ${}^\wedge$: the skew-symmetric matrix of a vector
    \item ${}^w \mathbf{p}_c$: camera position in world coordinate frame
    \item ${}^w_c \mathbf{R}\in SO(3)$: rotation matrix of camera in world coordinate frame
    \item $\boldsymbol\varphi^\wedge \in \mathfrak{se}(3)$: Lie-algebra representation of ${}^w_c \mathbf{R}$, where $exp(\boldsymbol\varphi^\wedge) = \mathbf{R}$ and $log(\mathbf{R}) = \boldsymbol\varphi^\wedge$
\end{itemize}

\subsection{DSO}
\label{sec:dso}
We choose DSO~\cite{engel2017direct} to implement visual odometry in our system. DSO works by minimizing the photometric error defined over a sliding window of keyframes (which are sparsely selected camera frames for optimization) with 3D points:

\begin{align}
    \label{eq:E_dso}
    E_{dso} = \sum_{j\in \mathcal{F}}\sum_{\mathbf{p}\in \mathcal{P}_j}\sum_{k\in obs(\mathbf{p})} e_{dso}(\mathbf{p},k) \\
    \label{eq:e_dso}
    \text{\footnotesize $e_{dso}(\mathbf{p},k) = \sum_{\mathbf{p}\in \mathcal{N}_\mathbf{p}} \mathit{w}_\mathbf{p} ||(I^k[\mathbf{p}']-b_k) - \frac{t_k e^{a_k}}{t_j e^{a_j}}(I^j[\mathbf{p}]-b_j)||_{\gamma}$}, \\
    \label{eq:dso_proj}
    \mathbf{p}' = \Pi({}_j^k\mathbf{T}\Pi^{-1}(\mathbf{p}, d_{\mathbf{p}})). 
\end{align}

For each point $\mathbf{p} \in \mathcal{P}_j$ in keyframe $j$ of sliding window $\mathcal{F}$, if $\mathbf{p}$ is observed by keyframe $k$, we calculate the photometric error $e_{dso}(\mathbf{p},k)$, which is essentially the pixel intensity ($I$ in Eq.~\ref{eq:e_dso}) difference between the point $\mathbf{p}$ in keyframe $j$ and its projection $\mathbf{p'}$ in keyframe $k$ as defined in Eq.~\ref{eq:dso_proj}. The affine brightness terms ($a_{j/k}$, $b_{j/k}$), exposure times ($t_{j/k}$), pixel pattern $\mathcal{N}_p$, and the Huber norm ($||\cdot||_{\gamma}$) in Eq.~\ref{eq:e_dso} are included for photometric robustness. Please refer to~\cite{engel2017direct} for more details.

\subsection{Spline Representation}
\label{sec:spline_intro}
A cubic spline~\cite{wikispline} is defined as 
\begin{align}
    \label{eq:cubic_spline}
    S(x) = C_j(x),\; x_{j-1} < x \leq x_j \\
    \label{eq:cubic_poly}
    C_j(x) = a_j + b_j x + c_j x^2 + d_j x^3 \\
    j = 1,...,n
\end{align}
with the boundary conditions of
\begin{align}
\label{eq:c0}
C_j(x_{j-1}) = y_{j-1},\; C_j(x_j) = y_j,\; j = 1,...,n \\
\label{eq:c1}
C_j'(x_j) = C_{j+1}'(x_j),\; j = 1,...,n-1 \\
\label{eq:c2}
C_j''(x_j) = C_{j+1}''(x_j),\; j = 1,...,n-1 
\end{align}
Intuitively, the cubic spline $S(x)$ is the continuous interpolation along a sequence of points $(x_1, y_1), ..., (x_n, y_n)$, with the constraints of zero/first/second-order continuities between segments $C_j$ and $C_{j+1}$. We refer to \cite{wikispline} for more details.

In the proposed VIO system, we maintain a sliding window of the keyframes: $KF_0, KF_1, KF_2, ... , KF_N$ where $N$ is the window size ($\leq 7$ in DSO by default). For $KF_j$, its camera pose is represented by ${}^w_{c_j} \mathbf{T} = [{}^w_{c_j} \mathbf{R}\; {}^w \mathbf{p}_{c_j}]$. We explicitly convert ${}^{c_j}_w \mathbf{T}$ in DSO to ${}^w_{c_j} \mathbf{T}$ in our system because positions represented in the uniform world coordinate frame are simple for spline representation (\ie Eq.~\ref{eq:pos}). We use a cubic spline to represent the continuous-time pose in the sliding window: the variable $x$ in Eq.~\ref{eq:cubic_spline} is now the time $t$, and $y$ is the keyframe pose $[{}^w_{c_j} \mathbf{R}\; {}^w \mathbf{p}_{c_j}]$; each spline segment $C_j$ represents the pose interpolation between two consecutive keyframes ($KF_j$ and $KF_{j-1}$ in the current implementation).

For position ${}^w \mathbf{p}_{c_j}$, since it is linear and represented in a uniform world coordinate frame, we directly interpolate it:
\begin{align}
\label{eq:pos}
{}^w \mathbf{p}_c(t) = {}^w \mathbf{p}_{c_j} + t \mathbf{l}^p_j + t^2 \mathbf{q}^p_j + t^3 \mathbf{c}^p_j \\
\label{eq:pos_cst}
{}^w \mathbf{p}_c(t_{j-1}) = {}^w \mathbf{p}_{c_{j-1}}
\end{align}
where $\mathbf{l}^p_j, \mathbf{q}^p_j, \mathbf{c}^p_j$ correspond to the \textbf{l}inear/\textbf{q}uadratic/\textbf{c}ubic coefficients $a_j, b_j, c_j$ in Eq.~\ref{eq:cubic_poly}. Unlike the global $x$ in Eq.~\ref{eq:cubic_spline}, the $t$ in Eq.~\ref{eq:pos} is the local time relative to $KF_j$; $t_{j-1}<0$ in Eq.~\ref{eq:pos_cst} represents the time of $KF_{j-1}$ with respect to $KF_j$.

For ${}^w_{c_j} \mathbf{R}\in SO(3)$, since it is not in a linear space, we cannot interpolate the rotation in world coordinate frame directly. Instead, we interpolate the relative rotation between ${}^w_{c_{j-1}} \mathbf{R}$ and ${}^w_{c_j} \mathbf{R}$:
\begin{align}
\label{eq:rot}
{}^w_c \mathbf{R}(t) = {}^w_{c_j} \mathbf{R} \cdot exp((t \mathbf{l}^r_j + t^2 \mathbf{q}^r_j + t^3 \mathbf{c}^r_j)^\wedge) \\
\label{eq:rot_cst}
{}^w_c \mathbf{R}(t_{j-1}) = {}^w_{c_{j-1}} \mathbf{R}
\end{align}

\subsection{IMU Synthesis}
If we take the second derivative of Eq.~\ref{eq:pos}, we get the camera acceleration in world coordinate frame:
\begin{equation*}
{}^w \mathbf{a}(t) = {}^w \mathbf{p}_c''(t) = 2 \mathbf{q}^p_j + 6t \mathbf{c}^p_j
\end{equation*}
However, IMU measures acceleration in local IMU coordinate frame, including the gravity; additionally, there is a bias term $\mathbf{b}^a_j$ associated to each keyframe $j$. Thus, we add the gravity $\mathbf{g}$, rotate it accordingly, and add the bias to synthesize IMU acceleration:
\begin{equation*}
{}^i \mathbf{a}(t) = {}^i_c \mathbf{R} \cdot {}^w_c \mathbf{R}(t)^T \cdot (2 \mathbf{q}^p_j + 6t \mathbf{c}^p_j + \mathbf{g}) + \mathbf{b}^a_j
\end{equation*}
There are two important notes: first, we explicitly estimate the scale $s$ of DSO in the metric world; second, we optionally estimate the roll $r$ and pitch $p$ angle of the gravity. The final synthesized acceleration data becomes:
\begin{align}
\label{eq:acc}
{}^i \mathbf{a}(t) = {}^i_c \mathbf{R} \cdot {}^w_c \mathbf{R}(t)^T \cdot [s \cdot (2 \mathbf{q}^p_j + 6t \mathbf{c}^p_j) + \mathbf{g}] + \mathbf{b}^a_j \\
\mathbf{g} = 9.8 \cdot [-sin(p)cos(r), sin(r), -cos(p)cos(r)]^T
\end{align}

Similarly, we take the first derivative of Eq.~\ref{eq:rot} and get the angular velocity in the camera coordinate frame:
\begin{equation*}
{}^c \boldsymbol\omega (t) = \mathbf{l}^r_j + 2t \mathbf{q}^r_j + 3t^2 \mathbf{c}^r_j
\end{equation*}
We rotate the angular velocity to the IMU coordinate frame and add the gyroscope bias $\mathbf{b}^g_j$ to synthesize gyroscope readings:
\begin{equation}
\label{eq:gyro}
{}^i \boldsymbol\omega (t) = {}^i_c \mathbf{R} \cdot (\mathbf{l}^r_j + 2t \mathbf{q}^r_j + 3t^2 \mathbf{c}^r_j) + \mathbf{b}^g_j
\end{equation}

\subsection{VIO State}
Consequently, our VIO state vector becomes
\begin{align*}
    \mathbf{state} = [f_x, f_y, c_x, c_y, s, r, p, \mathbf{kf}_1, \mathbf{kf}_2, ..., , \mathbf{kf}_N] \\
    \mathbf{kf}_j = [\mathbf{s}_{dso}, \mathbf{b}^a_j, \mathbf{b}^g_j, \mathbf{l}^r_j, \mathbf{q}^p_j, \mathbf{q}^r_j, \mathbf{c}^p_j, \mathbf{c}^r_j] \\
    \mathbf{s}_{dso} = [{}^w \mathbf{p}_{c_j}, {}^w \boldsymbol\varphi_{c_j}, a_j, b_j, d_{j1}, d_{j2}, ..., d_{jM}]
\end{align*}
For global variables, $f_x, f_y, c_x, c_y$ are the camera intrinsic parameters, $s$ is the DSO scale, and $r, p$ are the roll and pitch of the gravity. For $KF_j$, its state $\mathbf{kf}_j$ includes the following variables: $\mathbf{s}_{dso}$ is the DSO state, $\mathbf{b}^a_j, \mathbf{b}^g_j$ are IMU bias, and $\mathbf{l}^r_j, \mathbf{q}^p_j, \mathbf{q}^r_j, \mathbf{c}^p_j, \mathbf{c}^r_j$ are the cubic spline coefficients. Here we do not store $\mathbf{l}^p_j$ in our state because the linear velocity is not observable by either DSO or IMU (DSO estimates pose up to an unknown scale and IMU measures the acceleration). The $\mathbf{s}_{dso}$ contains the camera pose ${}^w \mathbf{p}_{c_j}, {}^w \boldsymbol\varphi_{c_j}$, the affine photometric parameters $a_j, b_j$, and the (inverse) depth $d_j$ of $M$ feature points associated with this keyframe.

\subsection{Spline Constraints}
Closer observation of Eq.~\ref{eq:acc} and Eq.~\ref{eq:gyro} shows that they are solely dependent on the current keyframe $KF_j$ and independent of $KF_{j-1}$. It is because we have not considered the spline boundary conditions (\ie Eq.~\ref{eq:c0}, Eq.~\ref{eq:c1}, Eq.~\ref{eq:c2}). Eq.~\ref{eq:rot} and Eq.~\ref{eq:rot_cst} connect ${}^w_{c_{j-1}} \mathbf{R}$ and ${}^w_{c_j} \mathbf{R}$ and introduce constraints to the rotation spline:
\begin{multline}
    \label{eq:c_r}
    \mathbf{c}_r(j) = {}^w_{c_{j-1}} \mathbf{R}^T \cdot {}^w_{c_j} \mathbf{R} \\
    \cdot exp((t_{j-1} \mathbf{l}^r_j + t_{j-1}^2 \mathbf{q}^r_j + t_{j-1}^3 \mathbf{c}^r_j)^\wedge) = \mathbf{I}
\end{multline}
Position constraints are more complex since we do not store $\mathbf{l}^p_j$ into our state vector. We use Eq.~\ref{eq:pos} and Eq.~\ref{eq:pos_cst} to calculate it:
\begin{equation}
\label{eq:vel_cal}
\mathbf{l}^p_j = ({}^w \mathbf{p}_{c_{j-1}} - {}^w \mathbf{p}_{c_j}) / t_{j-1} - t_{j-1} \mathbf{q}^p_j - t_{j-1}^2 \mathbf{c}^p_j
\end{equation}
Here we calculate the linear velocity $\mathbf{l}^p_j$ at $KF_j$ with the reference of $KF_{j-1}$. We can also predict it from the next keyframe $KF_{j+1}$:
\begin{multline*}
\mathbf{l}^p_j = {}^w \mathbf{p}^{*'}_c(t_j) = \mathbf{l}^p_{j+1} + 2t_j \mathbf{q}^p_{j+1} + 3t_j^2 \mathbf{c}^p_{j+1} \\
= ({}^w \mathbf{p}_{c_j} - {}^w \mathbf{p}_{c_{j+1}}) / t_j + t_j \mathbf{q}^p_{j+1} + 2t_j^2 \mathbf{c}^p_{j+1} 
\end{multline*}
${}^w \mathbf{p}^{*}_c$ is the spline segment starting from $KF_{j+1}$ (shift $j$ to $j+1$ in Eq.~\ref{eq:pos}). Making $\mathbf{l}^p_j$ consistent, we get the velocity continuity constraint:
\begin{multline}
    \label{eq:c_v}
    \mathbf{c}_v(j) = ({}^w \mathbf{p}_{c_{j-1}} - {}^w \mathbf{p}_{c_j}) / t_{j-1} - t_{j-1} \mathbf{q}^p_j - t_{j-1}^2 \mathbf{c}^p_j \\
     - ({}^w \mathbf{p}_{c_j} - {}^w \mathbf{p}_{c_{j+1}}) / t_j - t_j \mathbf{q}^p_{j+1} - 2t_j^2 \mathbf{c}^p_{j+1} = \mathbf{0}
\end{multline}

The rest of spline boundary conditions, such as linear acceleration continuity, are not explicitly constrained since they are implicitly constrained by the IMU measurements. Not constraining them also slightly relaxes the cubic spline motion assumption (\ie the motion is modeled by the cubic polynomial in Eq.~\ref{eq:cubic_poly}). In fact, constraining consecutive orientations (Eq.~\ref{eq:c_r}), positions (hardcoded into Eq.~\ref{eq:vel_cal}), and velocities (Eq.~\ref{eq:c_v}) is exactly what the IMU preintegration approach~\cite{forster2015imu} does.

\subsection{Energy Function}
\begin{figure}[ht]
    \centering
    \includegraphics[width=\textwidth]{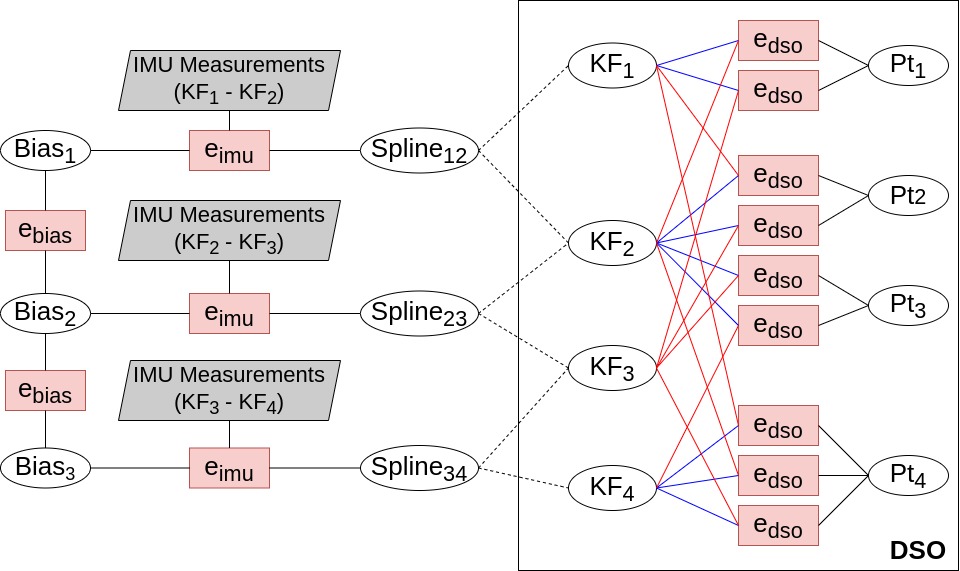}
    \caption{Factor graph for the proposed VIO system. Fig.~5 of \cite{engel2017direct} shows the details of the box labeled DSO. For the IMU part, $\mathbf{e}_{imu}$ is calculated by all IMU measurements between two consecutive keyframes and current IMU bias in Eq.~\ref{eq:e_imu}; $\mathbf{e}_{bias}$ is determined by consecutive IMU bias terms in Eq.~\ref{eq:e_bias}; the dashed lines between splines and keyframes (KF) are the spline constraints (Eq.~\ref{eq:c_r}, Eq.~\ref{eq:c_v}, Eq.~\ref{eq:energy}).}
    \label{fig:factor}
\end{figure}

\begin{figure}[ht]
    \centering
    \includegraphics[width=0.75\textwidth]{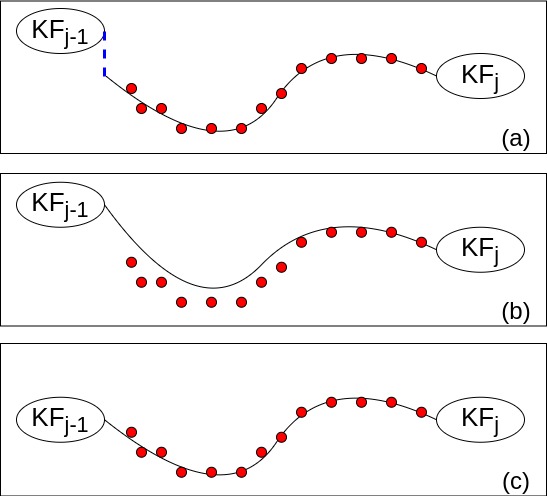}
    \caption{An intuitive explanation of the IMU part in our system. Between two consecutive keyframes ($KF_{j-1}$ and $KF_j$), we have many IMU measurements (red dots). The goal is to fit a spline (the curve between $KF_{j-1}$ and $KF_j$) to minimize the distance to the IMU measurements ($E_{imu}$) while the constraints ($\mathbf{c}_r=\mathbf{0}$ and $\mathbf{c}_v=\mathbf{0}$) are satisfied. (a): IMU error is minimized but the constraints are unsatisfied (the blue dashed line); (b): the constraints are satisfied but IMU error is large; (c): IMU error is minimized while the constraints are satisfied.}
    \label{fig:spline_explain}
\end{figure} 

Combining photometric error from DSO, IMU error between consecutive keyframes, and the spline constraints, we get our final energy function for optimization:
\begin{align}
    \label{eq:energy}
    min (E_{dso} + \lambda E_{imu})\; s.t.\; \mathbf{c}_r=\mathbf{I}\; and\; \mathbf{c}_v=\mathbf{0} \\
    E_{imu} = \mathbf{e}_{imu}^T \mathbf{W}_n \mathbf{e}_{imu} + \mathbf{e}_{bias}^T \mathbf{W}_b \mathbf{e}_{bias} \\
    \label{eq:e_imu}
    \mathbf{e}_{imu} = 
    \begin{bmatrix} {}^i \mathbf{a}(t) \\ \boldsymbol\omega(t) \end{bmatrix} 
    -
    \begin{bmatrix} {}^i \widetilde{\mathbf{a}}(t) \\ {}^i \widetilde{\boldsymbol\omega}(t) \end{bmatrix}\\
    \label{eq:e_bias}
    \mathbf{e}_{bias} = 
    \begin{bmatrix} \mathbf{b}^a_j \\ \mathbf{b}^g_j \end{bmatrix}
    -
    \begin{bmatrix} \mathbf{b}^a_{j-1} \\ \mathbf{b}^g_{j-1} \end{bmatrix}
\end{align}
$E_{dso}$ is already introduced in Sec.~\ref{sec:dso}. $E_{imu}$ is the sum of IMU measurement error $\mathbf{e}_{imu}$ weighted by IMU noise $\mathbf{W}_n$ and IMU bias error $\mathbf{e}_{bias}$ weighted by IMU random walk $\mathbf{W}_b$. $\mathbf{e}_{imu}$ is the difference from the predicted IMU reading by Eq.~\ref{eq:acc} and Eq.~\ref{eq:gyro} to IMU measurement $[{}^i \widetilde{\mathbf{a}}(t)^T,\; {}^i \widetilde{\boldsymbol\omega}(t)^T]^T$; $\mathbf{e}_{bias}$ is the bias difference between two consecutive keyframes. $E_{imu}$ is added to $E_{dso}$ with an empirical weight $\lambda$. The resulting factor graph is given in Fig.~\ref{fig:factor}. Fig.~\ref{fig:spline_explain} gives an intuitive explanation of our VIO system.
 
Since $\mathbf{c}_r$ and $\mathbf{c}_v$ are equality constraints, we can solve Eq.~\ref{eq:energy} efficiently using Lagrange multiplier~\cite{bertsekas2014constrained}.

\subsection{Initialization}
Using spline representation makes initialization more straightforward compared to the conventional approaches. First of all, the gravity is initialized using the first $40$ IMU measurements as \cite{von2018direct} does. When the system starts, we run DSO to get four keyframes. We assume these four keyframes share a set of spline coefficients (\ie $\mathbf{l}$, $\mathbf{q}$, $\mathbf{c}$ in Eq.~\ref{eq:pos} and Eq.~\ref{eq:rot}) so that we can linearly solve for the coefficients. After the spline is initialized, we can linearly solve for the IMU bias and scale using Eq.~\ref{eq:acc} and Eq.~\ref{eq:gyro} with the IMU measurements. Empirically, we find that setting acceleration bias $\mathbf{b}^a$ to zero ensures a more robust initialization process.

\subsection{First Estimate Jacobians}
\label{sec:method_discuss}
After the initialization stage, we use First Estimate Jacobians~\cite{huang2009first} throughout our implementation. Even though there is a large amount of IMU data, we only calculate the Jacobians once at the first estimate and store the value for later iterations. Using First Estimate Jacobians not only maintains the consistency of our VIO system \cite{engel2017direct} but also improves the computational efficiency.
\section{Experimental Evaluation}
\label{sec:experiments}
Through experiments, we show that the proposed VIO approach, as an alternative to IMU preintgration~\cite{forster2015imu}, achieves state-of-the-art performance. We refer to the proposed method as \textit{SplineVIO}. Since our VIO implementation is based on DSO~\cite{engel2017direct}, we compare it to DSO with IMU preintegration for direct comparison. There is no such official implementation; however, \cite{sun2021vidso} is a third-party implementation of VI-DSO~\cite{von2018direct} and Stereo-DSO~\cite{wang2017stereo}. Disabling the stereo component from \cite{wang2017stereo}, the remaining system is essentially VI-DSO, the state-of-the-art monocular VIO system of DSO with IMU preintegration using dynamic marginalization. Additionally, we include DSO as the baseline.

Our evaluation is based on the EuRoC dataset~\cite{burri2016euroc} and TUM VI dataset~\cite{schubert2018tumvi}. The EuRoC dataset is recorded by a Micro Aerial Vehicle (MAV) in two scenes: the `Machine Hall' (MH) and the `Vicon Room' (V), with the ground-truth poses available for the entire dataset. The TUM VI dataset is recorded in several scenes, however, only the `Room' (TR) sequences provide full ground-truth poses. Thus we focus on the `Room' sequences for a complete comparison. 

\begin{figure}[ht]
    \centering
    \includegraphics[width=\textwidth]{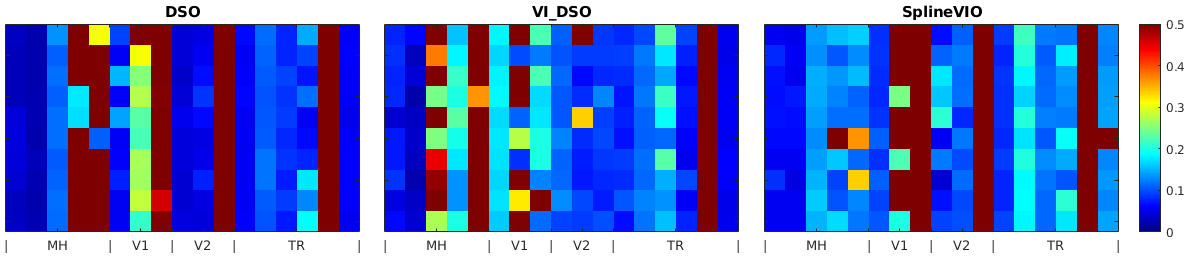}
    \caption{RMSEs (in meters) of $10$ runs (rows) for different methods on each sequence (columns) from EuRoC dataset (MH, V1, V2) and TUM VI dataset (TR). DSO results are aligned with $Sim(3)$; VI-DSO and SplineVIO are aligned with $SE(3)$.}
    \label{fig:color_plots}
\end{figure}

We mainly use the absolute trajectory error~\cite{engel2017direct, von2018direct} as the accuracy metric, which is calculated as the root mean square error (RMSE) of the trajectory. The trajectory is aligned to the ground-truth poses by $SE(3)$ for our system and VI-DSO, and by $Sim(3)$ for DSO since scale cannot be estimated by monocular visual odometry. Due to the randomness in DSO, we run each algorithm $10$ times and report the individual results in Fig.~\ref{fig:color_plots} and the median RMSE in Table~\ref{tab:rmses}.

\begin{table}[ht]
    \centering
    \caption{The median RMSEs over $10$ runs for different methods on each sequence from EuRoC dataset and TUM VI dataset. For VI-DSO, \cite{sun2021vidso} are the results of the 3rd party implementation; \cite{von2018direct} are the original results reported in the paper~\cite{von2018direct}, we include these results for reference.}
    \begin{subtable}{\textwidth}
    \centering
    \begin{tabular}{  l | l l l l l }  
    \Xhline{2\arrayrulewidth}
    Sequence & MH1 &  MH2 &  MH3 &  MH4 &  MH5 \\
    \Xhline{\arrayrulewidth}
    VI-DSO~\cite{sun2021vidso} & $0.076$ & $\mathbf{0.033}$ & $0.468$ & $0.190$ & $\times$ \\
    SplineVIO & $\mathbf{0.066}$ & $0.056$ & $\mathbf{0.142}$ & $\mathbf{0.131}$ & $\mathbf{0.129}$ \\
    \Xhline{\arrayrulewidth}
    VI-DSO~\cite{von2018direct} & $0.062$ & $0.044$ & $0.117$ & $0.132$ & $0.121$ \\
    DSO & $0.032$ & $0.024$ & $0.115$ & $\times$ & $0.929$ \\
    \Xhline{2\arrayrulewidth}
    \end{tabular}
    \caption{EuRoC Machine Hall}
    \label{tab:rmses_mh}
    \end{subtable}
    \vfill
    \begin{subtable}{\textwidth}
    \centering
    \begin{tabular}{  l | l l l l l l }  
    \Xhline{2\arrayrulewidth}
    Sequence & V11 & V12 & V13 & V21 & V22 & V23 \\
    \Xhline{\arrayrulewidth}
    VI-DSO~\cite{sun2021vidso} & $0.173$ & $\times$ & $\mathbf{0.187}$ & $0.110$ & $\mathbf{0.087}$ & $\mathbf{0.089}$ \\
    SplineVIO & $\mathbf{0.087}$ & $\times$ & $\times$ & $\mathbf{0.103}$ & $0.111$ & $\times$ \\
    \Xhline{\arrayrulewidth}
    VI-DSO~\cite{von2018direct} & $0.059$ & $0.067$ & $0.096$ & $0.040$ & $0.062$ & $0.174$ \\
    DSO & $0.059$ & $0.267$ & $0.590$ & $0.042$ & $0.053$ & $0.855$ \\
    \Xhline{2\arrayrulewidth}
    \end{tabular}
    \caption{EuRoC Vicon Room}
    \label{tab:rmses_v}
    \end{subtable}
    \vfill
    \begin{subtable}{\textwidth}
    \centering
    \begin{tabular}{  l | l l l l l l }  
    \Xhline{2\arrayrulewidth}
    Sequence & TR1 & TR2 & TR3 & TR4 & TR5 & TR6  \\
    \Xhline{\arrayrulewidth}
    VI-DSO~\cite{sun2021vidso} & $\mathbf{0.082}$ & $\mathbf{0.114}$ & $0.165$ & $\mathbf{0.072}$ & $\times$ & $\mathbf{0.055}$\\
    SplineVIO & $0.085$ & $0.186$ & $\mathbf{0.114}$ & $0.142$ & $\times$ & $0.137$\\
    \Xhline{\arrayrulewidth}
    VI-DSO~\cite{von2018direct} & - & - & - & - & - & -\\
    DSO & $0.058$ & $0.106$ & $0.084$ & $0.106$ & $\times$ & $0.059$\\
    \Xhline{2\arrayrulewidth}
    \end{tabular}
    \caption{TUM VI Room}
    \label{tab:rmses_tr}
    \end{subtable}
    \label{tab:rmses}
\end{table}

\subsection*{EuRoC Dataset}

\begin{wrapfigure}{r}{0.5\textwidth}
    \centering
    \includegraphics[width=0.4\textwidth]{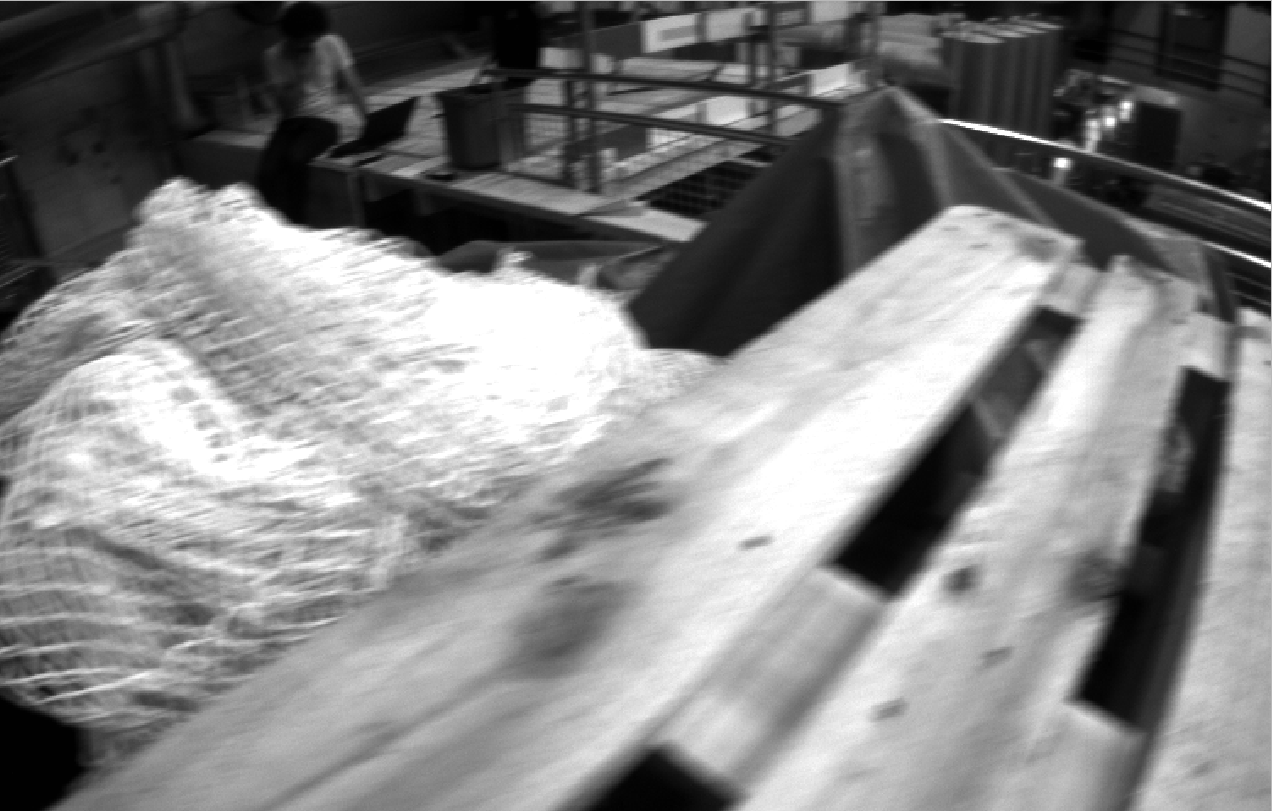}
    \caption{The MH sequences start with fast motion to activate IMU, which causes motion blur that creates challenges for visual tracking.}
    \label{fig:mh_init}
\end{wrapfigure}

\paragraph*{Machine Hall} Fast motion is present at the beginning of each MH sequence to aid IMU initialization, but the resulting motion blur is challenging for visual tracking. An example is given in Fig.~\ref{fig:mh_init}. That is why the authors of DSO~\cite{engel2017direct} crop the beginning and only use the part of the sequence where the MAV is in the air. However, for VIO systems, IMU initialization is essential so we cannot discard the beginning part. Hence, our experiments use the entire sequence. Parameter-wise, we set the IMU weight $\lambda=6$ in Eq.~\ref{eq:energy}, which is the default value in \cite{sun2021vidso} for testing on EuRoC dataset.

From the MH parts in Fig.~\ref{fig:color_plots} and Table~\ref{tab:rmses_mh}, DSO works on MH1 to MH3 but fails on MH4 and MH5. The results of DSO on MH sequences are slightly worse than the results reported in \cite{engel2017direct} since we do not crop the beginning parts. Similarly, our results of VI-DSO on MH sequences are not as good as the results in \cite{von2018direct} (we include the original results to the row of VI-DSO~\cite{von2018direct} in Table~\ref{tab:rmses} for reference). The first reason is that we use a third-party implementation~\cite{sun2021vidso} since no official release is available; the other reason is that the authors of \cite{von2018direct} do not use the entire sequence and we have no information about their start time. Compared to DSO, the results of VI-DSO on MH3 are worse but the results on MH4 are better. This is because well initialized and maintained IMU states improve the overall accuracy and robustness while poorly initialized or maintained IMU states degrade performance. The proposed SplineVIO performs the best on MH sequences, it works most of the time on all MH sequences (Fig.~\ref{fig:color_plots}) and achieves lower median RMSEs in Table~\ref{tab:rmses_mh}.

\paragraph*{Vicon Room} The Vicon Room sequences do not contain fast motion for IMU initialization. Hence, the DSO results are closer to the ones reported in \cite{engel2017direct}. However, due to the lack of sufficient IMU initialization, the performance of SplineVIO degrades compared to MH sequences. VI-DSO is more robust and slightly more accurate on the Vicon Room sequences compared to SplineVIO. Dynamic marginalization in VI-DSO enables delayed IMU initialization that is very effective on sequences without fast motion at the beginning such as Vicon Room sequences. We intend to integrate dynamic marginalization into our system in future work for improved accuracy and robustness. Nevertheless, the accuracy margin between VI-DSO and SplineVIO in Table~\ref{tab:rmses_v} is not significant on sequences that both work (\ie V11, V21, V22). 

\subsection*{TUM VI Dataset}
To further validate the proposed SplineVIO, we test it on the Room sequences in the TUM VI dataset~\cite{schubert2018tumvi}. Due to the different hardware setup, we change the IMU weight $\lambda=0.1$ in Eq.~\ref{eq:energy} (same for the IMU weight in VI-DSO). Similar to the Vicon Room sequences, VI-DSO slightly outperforms the proposed SplineVIO but the margin is small. This dataset does not have a dedicated IMU initialization stage, the dynamic marginalization in VI-DSO helps IMU initialization and improves the overall accuracy.

The evaluations on the EuRoC dataset and TUM VI dataset show that the proposed SplineVIO achieves comparable accuracy compared to the state-of-the-art DSO-based VIO system (\ie VI-DSO). 

\subsection*{Computational Efficiency}
\begin{table}[ht]
    \centering
    \caption{The average run-time on keyframe optimization and overall FPS over 10 runs.}
    \begin{tabular}{l|l l l}
        \Xhline{\arrayrulewidth}
        Method & DSO & VI-DSO & SplineVIO \\
        \Xhline{\arrayrulewidth}
        KF opt. (ms) & $22.9$ & $24.9$ & $34.2$ \\
        FPS & $65.1$ & $55.2$ & $55.8$ \\
        \Xhline{\arrayrulewidth}
    \end{tabular}
\end{table}

We run each algorithm on the MH1 sequence $10$ times and take the average time for efficiency evaluation. The keyframe optimization time for VI-DSO does not increase significantly over DSO because IMU measurements are only integrated once and then serve as a unit measurement using IMU preintegration. For SplineVIO, the keyframe optimization is slower. Even though we do not need to recalculate Jacobians for each iteration as discussed in Sec.~\ref{sec:method_discuss}, we do have to recalculate the IMU errors using Eq.~\ref{eq:acc}, Eq.~\ref{eq:gyro}, and Eq.~\ref{eq:e_imu}. Given the high volume of IMU measurement ($200Hz$), the IMU error calculation is computationally expensive. One note is that we have not intensively optimized the implementation for efficiency. Currently, the IMU error calculation is done one by one in a loop, which can be significantly improved by using single-instruction-multiple-data (SIMD) technology as DSO does. Nevertheless, $34.2ms$ for keyframe optimization is still acceptable since keyframe optimization only runs when a new keyframe is created. 

As for the overall FPS, SplineVIO is marginally higher than VI-DSO. VI-DSO puts the IMU error terms into the visual tracking energy function, which improves the tracking accuracy and robustness by sacrificing some computational efficiency. For SplineVIO, we use the spline to predict a pose for visual tracking and the IMU error terms are not used in visual tracking. With a faster back-end (keyframe optimization) but a slower front-end (IMU involved visual tracking), the FPS of VI-DSO is thus not higher than that of SplineVIO. Nevertheless, we show that the proposed SplineVIO system is capable of running in real-time.
\section{Conclusions}
In this paper, we propose a VIO approach based on spline. Specifically, we use a cubic spline to represent the continuous-time pose in the DSO keyframe window. We use the temporal derivatives of the spline to synthesize IMU measurements and calculate IMU errors. With the spline boundary conditions, we formulate VIO as a constrained nonlinear optimization problem. Through experiments on the EuRoC dataset and TUM VI dataset, we show that the proposed VIO system achieves state-of-the-art accuracy and real-time efficiency. For future work, we plan to optimize the implementation to further improve accuracy, robustness, and efficiency. We also plan to extend this work for rolling shutter cameras using the benefit of continuous-time spline representation.

\section*{Acknowledgment}
This work was partially supported by the Minnesota Robotics Institute Seed (MnRI) Grant and the National Science Foundation award IIS-\#1637875.

\bibliographystyle{ieeetr}
\bibliography{main}
\end{document}